\documentclass[letterpaper, 10 pt, conference]{ieeeconf}  

\IEEEoverridecommandlockouts                              
\usepackage{amsmath}
\usepackage{graphicx}
\usepackage{booktabs}
\usepackage{makecell}
\usepackage{tabularx}
\usepackage{xcolor}
\usepackage{float}
\usepackage{cite}        
\usepackage{url}
\usepackage[hidelinks]{hyperref}
\usepackage[most]{tcolorbox}

\definecolor{seabornblue}{RGB}{31,119,180}
\newcounter{takeawaycounter}
\newenvironment{takeaway}{
  \stepcounter{takeawaycounter}
  \begin{tcolorbox}[
    colback=seabornblue!10,
    colframe=black,
    boxrule=0.5pt,
    arc=4pt,
    width=\linewidth,
    left=6pt, right=6pt, top=4pt, bottom=4pt,
  ]
  \textbf{Takeaway~\thetakeawaycounter:}
}{
  \end{tcolorbox}
}

\title{\LARGE \bf
MolmoSpaces-CommonSense: Evaluating Commonsense in Robotic Manipulation Policies
}

\author{Arjun Guru$^{1,2}$, Max Argus$^{2}$, Jae Sung Park$^{2}$, Jiafei Duan$^{3}$,
        Ali Farhadi$^{1}$, and Ranjay Krishna$^{1}$
\thanks{$^{1}$Paul G. Allen School of Computer Science \& Engineering,
        University of Washington, Seattle, WA, USA}%
\thanks{$^{2}$Allen Institute for AI (Ai2), Seattle, WA, USA}%
\thanks{$^{3}$National University of Singapore, Singapore}%
}

\begin{document}

\bstctlcite{BSTcontrol}

\maketitle
\thispagestyle{empty}
\pagestyle{empty}

\begin{abstract}
Real-world robotic manipulation requires reasoning about cluttered scenes, partially observable or hidden objects, and semantically appropriate actions, in addition to low-level execution within those scenes. However, existing simulation benchmarks primarily test the latter and remain dominated by comparatively clean environments with limited clutter and narrowly scoped task structure, leaving many forms of commonsense reasoning untested. We address this by introducing a taxonomy of reasoning processes for robotic manipulation grounded in developmental cognitive science. We operationalize this taxonomy through a new benchmark of robot tasks, \textit{MolmoSpaces-CommonSense}, that requires reasoning about occlusion, containment, support relations, semantically correct grasp points, spatial affordances, and hidden objects. These evaluations require policies not only to recognize relevant scene structure but also to execute correct actions in them. We evaluate several zero-shot generalist manipulation policies. Across model types and architectures, we observe consistent failures in commonsense reasoning: success rates in cluttered scenes drop by over 50\% as distractors increase, policies act on incorrect objects in cluttered settings 30\% of the time, and affordance-sensitive grasping remains unreliable even with highly specific language prompts, with correct object part selection happening only about 50\% of the time. 
Further, VLM-based success estimation struggles to reliably evaluate rollouts in these settings, suggesting that the challenge extends beyond control to the broader perception-reasoning-action stack. 
Our results suggest that existing benchmarks may substantially overestimate zero-shot real-world readiness. Code and policy leaderboard link at https://github.com/allenai/molmospaces.
\end{abstract}


\section{Introduction}

Manipulation in the real world is a reasoning problem before it is a control problem. A robot must decide which object an instruction refers to in clutter, infer what remains true when objects become partially hidden, determine what actions the scene affords, and choose behaviors that are physically and semantically appropriate for the goal. Cognitive science has long treated these capabilities not as a single faculty but as interacting forms of physical reasoning, affordance perception, and object permanence~\cite{baillargeon2004,gibson1979,piaget1954}. Yet the benchmarks that currently anchor progress in robot learning~\cite{black__0_2024,pertsch_fast_2025,intelligence__05_2025,team_gemini_2025-1,team_gemini_2025} are still built primarily around executing predefined task suites in comparatively controlled scenes with a small set of task-relevant objects, rather than systematically requiring reasoning under clutter, occlusion, or hidden state. 

\begin{figure*}[t]
\centering
\includegraphics[width=.98\textwidth]{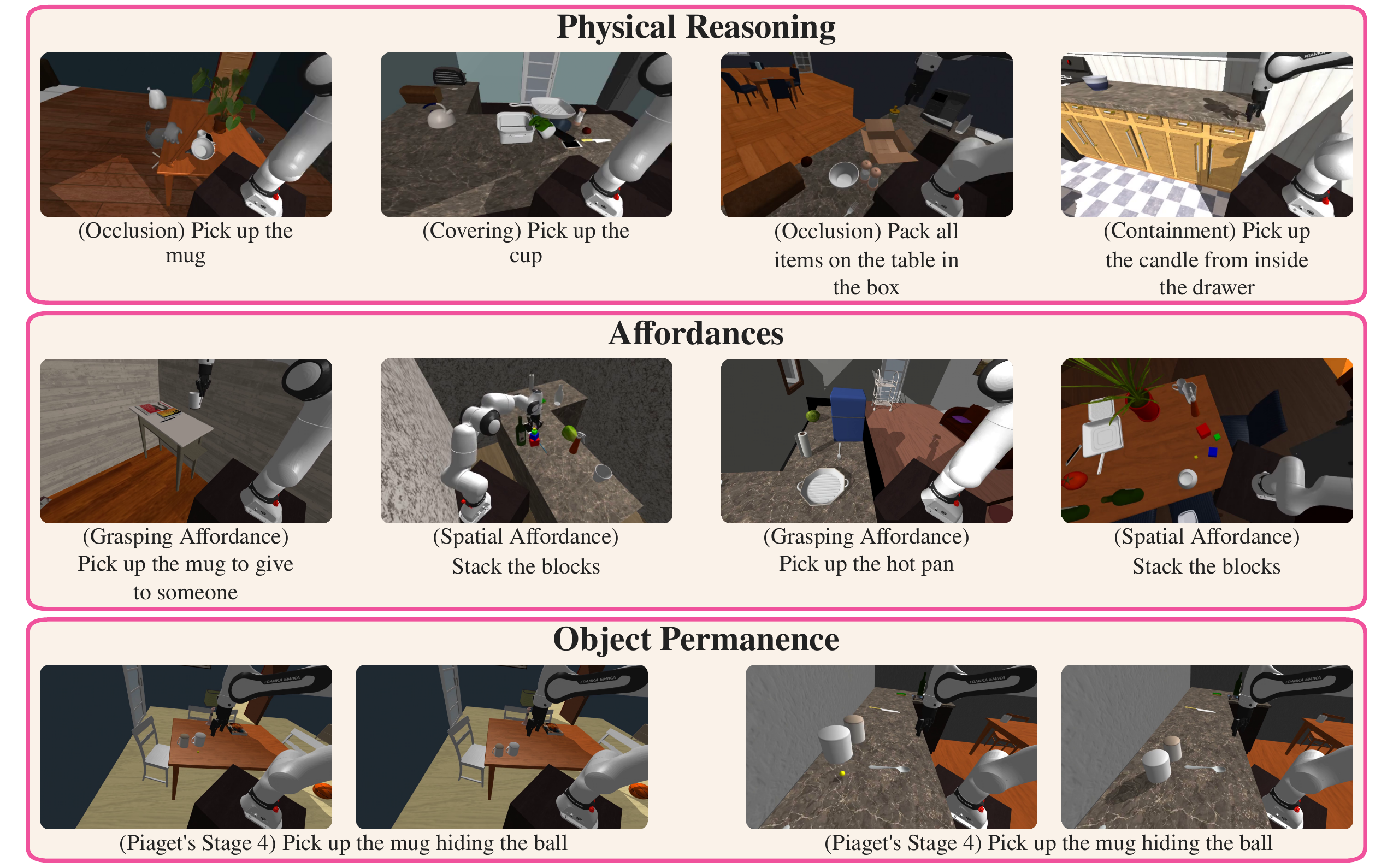}
\caption{Taxonomy of embodied commonsense reasoning capabilities, showing the diversity of tasks we test on. We organize tasks along three categories grounded in developmental cognitive science, and instantiate tasks within MolmoSpaces environments.}
\vspace{-1.5em}
\label{fig:taxonomy}
\end{figure*}

This gap is ubiquitous in homes, labs, and workspaces, which are cluttered and where targets are often occluded, covered, contained, or displaced out of view. Yet current robot policies are largely neither trained nor evaluated in environments that make these capabilities necessary~\cite{kim_molmospaces_2026}: a policy that succeeds only when the target is cleanly visible and the correct action is obvious is not one that can be deployed in the real world. 

In this work, we introduce a taxonomy of commonsense reasoning processes for robotic manipulation, grounded in decades of cognitive science. The taxonomy (see Fig.~\ref{fig:taxonomy}) organizes a practically important subset of embodied commonsense into three categories: \textit{physical reasoning} (object presence and relations under occlusion, covering, and containment), \textit{affordances} (grasping affordances at the object-part level and spatial affordances at the surface level), and \textit{object permanence} (acting on objects seen but no longer visible). Our goal is not to claim completeness but to provide a concrete framework for asking which reasoning processes current manipulation policies actually utilize.

We operationalize this taxonomy by extending MolmoSpaces~\cite{kim_molmospaces_2026} with environments that force these reasoning processes to matter at action time. MolmoSpaces provides a scalable ecosystem of diverse indoor scenes, richly annotated objects, and infrastructure with strong sim-to-real alignment, making it a natural substrate for this evaluation. On top of it, we construct tasks and conditions that demand reasoning at action time, such as dense distractor clutter, targets hidden beneath covers or inside containers, and prompts requiring semantically correct grasp points. Because these evaluations are closed-loop, success requires more than recognizing the right relation in an image; the policy must use that relation to produce the right action sequence. 

We use MolmoSpaces-CommonSense to evaluate representative zero-shot generalist policies spanning both VLA and WAM paradigms, including $\pi_{0.5}$~\cite{intelligence__05_2025}, MolmoAct2~\cite{molmoact2_2026}, and DreamZero~\cite{dreamzero2026}. The main pattern is strikingly consistent across architectures. In cluttered physical-reasoning tasks, success drops by more than 50\% as the number of nearby distractors increases, and the dominant failure mode is not generic instability but selecting or moving toward the wrong object. In affordance-sensitive grasping tasks, overall success remains low, around 25\%, and increasing prompt specificity does little to improve whether the object is grasped the correct part. Together, these results suggest that current policies can often execute atomic motions once the scene is easy, but they do not yet reliably reason through the scene conditions that make real manipulation hard. We further find that VLM judges struggle to reliably classify success and failure from rollout videos in these scenes, suggesting that the difficulty extends across the broader perception-reasoning-action stack.

More broadly, this study is about what we choose to measure. If benchmarks do not test hidden-object reasoning, affordance-sensitive action selection, or behavior under partial observability, then high success rates can tell an overly optimistic story about readiness for deployment. By grounding benchmark design in cognitive science and instantiating that design in scalable embodied environments, we hope to make commonsense reasoning a first-class target for robot policy evaluation and improvement.
\section{Related work}
\label{sec:related}

Evaluating the reasoning capabilities of generalist robot policies draws on three lines of work: commonsense reasoning evaluation in foundation models, embodied reasoning benchmarks for robot policies, and simulation environments that make such evaluations reproducible and scalable. 

\noindent\textbf{Commonsense Reasoning Evaluation of Foundation Models.}
Commonsense reasoning has been studied extensively in language and vision-language models~\cite{tur2026recurrent,sap_atomic_2019,zellers_swag_2018,zellers_hellaswag_2019,sakaguchi_winogrande_2019,park_visualcomet_2020,hessel_abduction_2022,duan2022pip,hudson_gqa_2019}, often via taxonomy-based or adversarially-filtered benchmarks, which inspire our taxonomy-driven evaluation of robot policies. Concerning physical and embodied reasoning, IntPhys~\cite{riochet_intphys_2020} focuses on evaluating intuitive physics understanding, and Embodied Reasoning Question Answering~\cite{team_gemini_2025-1} evaluates reasoning about actions, trajectories, 3D relationships, affordances, and more. Though a significant step, the goal of assessing embodied agents' reasoning capabilities ultimately requires closing the loop between reasoning and action and evaluating models' capabilities while they interact with the world. Our work aims to bridge this gap and test model capabilities in embodied settings beyond visual question-answering.

\noindent\textbf{Embodied Reasoning Evaluations.}
As generalist manipulation models improve at atomic actions, a number of works have begun evaluating their higher-level reasoning across different subsets of embodied reasoning tasks. Works such as Colosseum~\cite{pumacay_colosseum_2024} primarily study perturbations to tasks in simulation along multiple axes and measure policy robustness. STAR-Gen~\cite{gao_taxonomy_2026} develops a taxonomy on visual, semantic, and behavioral generalization and tests policies in the real world. VLABench~\cite{zhang_vlabench_2024} tests a categorization of VLA capabilities spanning visual, language, and physical understanding. Additional targeted benchmarks for specific components of embodied reasoning, such as memory \cite{dai_robomme_2026,fang_sam2act_2025}, affordance~\cite{yuan2024robopoint, cheng2025pointarena, hao_roboafford_2025}, and spatial reasoning~\cite{chen_spatialvlm_2024, song_robospatial_2026} have been developed. However, existing taxonomies often omit notable dimensions of human commonsense reasoning capabilities, whereas capability-specific benchmarks only evaluate certain dimensions in isolation. Moreover, concurrent reasoning benchmarks~\cite{lin2026robowits,chen2026volo,bhat2026probe,maurya2026imbench,chen2026robodojo,wang2026coin} are each confined to a tabletop workspace, varying object placement and in some cases textures and lighting; MolmoSpaces-CommonSense instantiates tasks across 1,009 distinct indoor scenes and 413 objects with randomized camera and base poses. Our work aims to build upon past efforts and provide a taxonomy grounded in cognitive science literature on human embodied commonsense reasoning capabilities, along with a set of tasks in simulation that allow for reproducible and scalable evaluations of these capabilities in diverse settings.

\noindent\textbf{Simulation benchmarks for robot policies.}
Simulation-based benchmarks have gained adoption as a scalable, reproducible alternative to real-world evaluation for robot manipulation~\cite{james_rlbench_2019,liu_libero_2023,mees_calvin_2022}. More recent benchmarks have expanded to include language grounding and generalization capabilities \cite{liu_libero_2023, mees_calvin_2022, fei_libero-plus_2025, zhou_libero-pro_2025, nasiriany_robocasa_2024, nasiriany_robocasa365_2026}, robustness to perturbations~\cite{pumacay_colosseum_2024}, a more extensive set of short- and long-horizon tasks within the household~\cite{li_behavior-1k_2024}, and alignment with the real world~\cite{li_evaluating_2024}. A number of these benchmarks require training in simulation~\cite{liu_libero_2023, mees_calvin_2022, li_behavior-1k_2024, nasiriany_robocasa_2024}, limiting their practicality in evaluating zero-shot policies. In contrast, MolmoSpaces~\cite{kim_molmospaces_2026} supports zero-shot evaluations of generalist robot policies in over 20K indoor scenes and over 22K interactable articulated and rigid objects~\cite{kolve_ai2-thor_2022, deitke_procthor_2022, yang_holodeck_2024}. Our work builds upon the MolmoSpaces ecosystem while extending the list of tasks beyond solely atomic actions to reflect our taxonomy on commonsense reasoning.

\section{Methods}
\label{sec:methods}

\begin{figure*}[t]
\centering
\includegraphics[width=\textwidth]{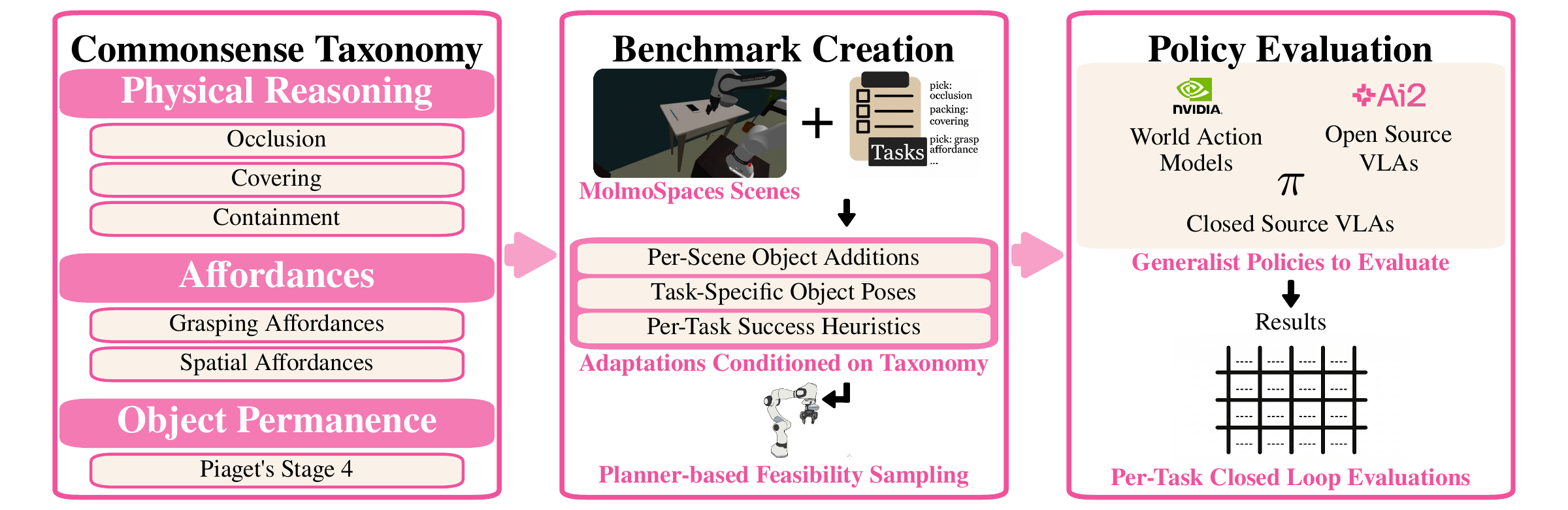}
\caption{Our systematic process for analyzing commonsense capabilities of models. Starting from a taxonomy of commonsense tasks, we implement corresponding benchmarks by utilizing the Molmospaces ecosystem and evaluate and analyze diverse policy classes.}
\vspace{-1em}
\label{fig:method}
\end{figure*}

Guided by our cognitive science-grounded task taxonomy, we design MolmoSpaces-CommonSense, a benchmark testing embodied commonsense reasoning and use it to evaluate generalist manipulation policies. Fig.~\ref{fig:method} illustrates our approach.

\subsection{Commonsense Taxonomy}
\label{sec:taxonomy}

Inspired by seminal studies in cognitive science on capabilities that emerge early in childhood development, we define a taxonomy that captures a meaningful subset of commonsense reasoning capabilities useful in manipulation tasks. We consider three categories of commonsense reasoning capabilities: \textit{physical reasoning}, \textit{affordances}, and \textit{object permanence}. Fig.~\ref{fig:taxonomy} illustrates our taxonomy and representative tasks per category. 

\textit{Physical reasoning} refers to reasoning about object presence and relationships under partial observability, inspired by Renée Baillargeon's work on three key infant event categories~\cite{baillargeon2004}: \textit{occlusion}, \textit{covering}, and \textit{containment}. A target object may be surrounded by occluding clutter (occlusion), hidden beneath covering object(s) (covering), or enclosed within a container such as a drawer (containment). While Baillargeon's paradigms probe perceptual expectations about hidden objects, we consider a closed-loop control setting, where the agent must act—not merely predict—under each partial observability condition and reason about avoiding or moving aside distractors to manipulate the target objects.

Next, we consider \textit{affordances}, inspired by James J. Gibson's notion~\cite{gibson1979} of the actions that an object or environment permits. We consider two key subcategories of affordances. We use the term \textit{spatial affordances} to refer to action possibilities stemming from surface layouts and object geometries; in a block-stacking task, for example, a block only affords supporting blocks its size or smaller, so a policy must reason about stacking blocks in descending order of size. We use \textit{grasping affordances} to refer to an expansion of Gibson's notion within robotics: the actions that specific object parts facilitate~\cite{myers2015affordance, do2018affordancenet}. For example, a mug affords grasping by its handle when it must be handed to someone (to avoid touching the part they will drink from). We operationalize affordance reasoning through tasks requiring policies to identify action-relevant object parts or spatial configurations and execute corresponding manipulations. 

Finally, we consider \textit{object permanence}, inspired by Jean Piaget's studies~\cite{piaget1954} on children's ability to interact with objects not in view but previously observed or interacted with. Piaget identified multiple stages of object permanence, ranging in difficulty from finding objects after observing their occlusion to tracking those objects through displacements while occluded. We develop tasks testing Piaget's Stage 4, in which an agent must be capable of recovering an object after observing its occlusion in one of multiple possible locations without tracking displacements.

\subsection{MolmoSpaces-CommonSense Benchmark Construction}
\label{sec:environments}

\begin{table*}[t]
\centering
\caption{Overview of the tasks in our benchmark, MolmoSpaces-CommonSense.}
\label{tab:tasks}
\renewcommand{\arraystretch}{1.15}
\setlength{\tabcolsep}{4pt}
\small
\begin{tabularx}{\textwidth}{@{}
  >{\raggedright\arraybackslash}X
  >{\raggedright\arraybackslash}p{2.0cm}
  >{\raggedright\arraybackslash}p{2.0cm}
  c
  >{\raggedright\arraybackslash}X @{}}
\toprule
\textbf{Task} & \textbf{Category} & \textbf{Subcategory} & \textbf{\# Envs} & \textbf{Success Condition} \\
\midrule
Pick with Occlusion    & Phys. Reas. & Occlusion   & 1000 & Target object lifted \\
Pick with Covering     & Phys. Reas. & Covering    & 1000 & Target object lifted \\
Pick with Containment  & Phys. Reas. & Containment & 500  & Target object lifted \\
Pack with Occlusion    & Phys. Reas. & Occlusion   & 1000 & All objects in receptacle \\
Pack with Covering     & Phys. Reas. & Covering    & 1000 & All objects in receptacle \\
Pick with Affordance        & Affordances     & Grasping    & 500  & Lifted + correct grasp pose \\
Stack with Affordance    & Affordances     & Spatial     & 1000 & All blocks stacked \\
Pick with Object Permanence  & Object Per.  & Piaget St.\ 4 & 1000 & Correct mug will ball underneath lifted \\
\bottomrule
\end{tabularx}
\end{table*}

We instantiate our taxonomy as a set of eight manipulation tasks distributed across the above categories (see Table~\ref{tab:tasks}), implemented within the MolmoSpaces ecosystem, which uses the MuJoCo physics simulator. Tasks were selected to be tractable to simulate and reflect realistic manipulation settings. For all tasks, initial scenes per environment are sampled from the MSCrafted~\cite{kolve_ai2-thor_2022} and MSProc~\cite{deitke_procthor_2022} scene datasets, and we randomize target objects, object poses, exocentric camera poses, and robot base poses. All environments use a DROID~\cite{khazatsky_droid_2025} setup (i.e., a Franka FR3 with the RobotIQ 2F-85 gripper, along with matching camera intrinsics and extrinsics). We additionally apply a planner-based feasibility filter during procedural generation: candidate environments are retained only if a privileged task-specific planner succeeds (following MolmoBot's~\cite{molmobot2026} data generation pipeline; see Appendix for more details).

\textbf{Physical reasoning.} We instantiate single-object pick and packing tasks for each subcategory. Environment creation entails procedural repositioning of distractor objects: surrounding the target (occlusion), covering it (covering), or enclosing it within an articulated object (containment).

\textbf{Affordances.} We instantiate one task per subcategory. For grasping affordances, we design pick tasks with targeted prompts, e.g.\ ``pick up the hot pan'', where success requires not only lifting the object but grasping it at the task-appropriate part (e.g.,\ the pan's handle). For spatial affordances, we design a block-stacking task (with randomized placement of 2-4 blocks in each environment) in which the policy must reason about larger blocks' ability to support smaller ones.

\textbf{Object Permanence.} We instantiate a single task targeting Piaget's Stage 4: each environment initializes two upside-down mugs suspended above a surface with a ball under one, providing a momentary observation window before the mugs fall onto the surface and hide the ball. The policy must subsequently pick the correct mug to reveal the ball.

\subsection{Evaluation Protocol}
\label{sec:eval}

\textbf{Robot Policy Evaluations.} We evaluate policies closed-loop and report per-task success rates with Bayesian beta-binomial credible intervals. Success is checked programmatically in MuJoCo (lift height, receptacle containment via ray-casting, grasp-pose matching against annotations, and block-contact checks); see Appendix for full specifications.

\textbf{VLM as a Judge Evaluations.} We additionally evaluate how well VLMs can \emph{verify} policy rollouts---a strictly easier task than \emph{generating} action trajectories, since the verifier only needs to recognize whether the trajectory succeeded, not produce one. This partially isolates perception and reasoning from action. For the Pick with Occlusion task, we annotate a sample of rollouts from representative policies with one of five outcome categories (four failure modes plus success; see Appendix), then prompt several VLMs to classify each rollout video and measure accuracy against these annotations.

\section{Experimental Results}
\label{sec:result}

To begin, we report success rates of various generalist policies when evaluated zero-shot on all tasks within MolmoSpaces-Commonsense (see Table~\ref{tab:headline_results}). Our evaluation spans policies with diverse architectures: $\pi_{0.5}$, $\pi_{0}$-FAST, and $\pi_{0}$ (all VLAs), MolmoAct2 (VLA with depth token reasoning traces), and DreamZero (WAM). Notably, no policy achieves a success rate over 30\% on any of our tasks. Our distributional analyses further reveal that \textit{these architecturally distinct policies fail in highly correlated ways}, sharing not only success rates but dominant failure modes. Since all evaluated policies are finetuned on DROID, where the conditions our tasks require are rare, we hypothesize that the bottleneck lies in the shared finetuning data rather than in any particular architecture (Sec.~F).

More broadly, we find that across all taxonomy categories, policies fall well short of capable performance. Within physical reasoning tasks, success drops below 20\% under occlusion or covering and to 0\% under containment. Policies struggle to recognize and act upon grasping and spatial affordances (with success on the latter being near zero). Additionally, no policy succeeds on object permanence due to inability to handle observation history.

We scope distributional analyses to tasks with non-zero zero-shot performance. Containment is excluded as uninformative and plausibly out-of-distribution, and object permanence is excluded due to the above structural concerns with existing generalist policies. We leave both to future work. Our VLM evaluations use Gemini 3.1~\cite{gemini31pro_2026}, Gemini Robotics ER 1.6~\cite{geminiroboticser16_2026}, Qwen3.5~\cite{qwen35_2026}, and Molmo2~\cite{clark2026molmo2openweightsdata}.

\newcommand{\val}[2]{$#1_{\pm #2}$}

\begin{table*}[t]
\centering
\small
\caption{Success rates (\%) across policies and tasks, showing policies struggle across all tasks.}
\label{tab:headline_results}
\setlength{\tabcolsep}{4pt}
\resizebox{\textwidth}{!}{%
\begin{tabular}{lcccccccc}
\toprule
 & \multicolumn{5}{c}{\textit{Physical Reasoning}} & \multicolumn{2}{c}{\textit{Affordance}} & \textit{Obj. Perm.} \\
\cmidrule(lr){2-6} \cmidrule(lr){7-8} \cmidrule(lr){9-9}
Policy & \makecell{Pick w/\\Occlusion} & \makecell{Pick w/\\Covering} & \makecell{Pick w/\\Containment} & \makecell{Pack w/\\Occlusion} & \makecell{Pack w/\\Covering} & \makecell{Pick w/\\Affordance} & \makecell{Stack w/\\Affordance} & \makecell{Pick w/ Obj.\\Permanence} \\
\midrule
$\pi_0$      & \val{2.81}{1.22}  & \val{2.80}{1.22}  & \val{0.00}{0.73} & \val{3.30}{1.30}  & \val{4.30}{1.44}  & \val{9.40}{2.88}  & \val{0.10}{0.46} & \val{0.00}{0.37} \\
$\pi_0$-FAST & \val{8.21}{1.87}  & \val{6.91}{1.74}  & \val{0.00}{0.73} & \val{2.20}{1.11}  & \val{5.60}{1.60}  & \val{13.40}{3.27} & $\boldsymbol{0.20_{\pm 0.52}}$ & \val{0.00}{0.37} \\
$\pi_{0.5}$  & \val{14.10}{2.30} & \val{11.31}{2.02} & \val{0.00}{0.73} & $\boldsymbol{12.10_{\pm 2.17}}$ & $\boldsymbol{14.20_{\pm 2.30}}$ & \val{19.40}{3.70} & $\boldsymbol{0.20_{\pm 0.52}}$ & $\boldsymbol{0.30_{\pm 0.57}}$ \\
DreamZero    & $\boldsymbol{18.00_{\pm 2.50}}$ & \val{13.80}{2.28} & \val{0.00}{0.73} & \val{7.00}{1.75}  & \val{7.10}{1.76}  & $\boldsymbol{26.60_{\pm 4.04}}$ & \val{0.10}{0.46} & $\boldsymbol{0.30_{\pm 0.57}}$ \\
MolmoAct2    & \val{17.20}{2.47} & $\boldsymbol{14.00_{\pm 2.29}}$ & \val{0.00}{0.73} & \val{8.96}{1.94}  & \val{10.14}{2.03} & \val{23.60}{3.92} & $\boldsymbol{0.20_{\pm 0.52}}$ & \val{0.00}{0.37} \\
\midrule
Planner      & 100.00            & 100.00            & 100.00           & 100.00            & 100.00            & 100.00            & 100.00           & 100.00 \\
\bottomrule
\end{tabular}
}
\end{table*}

\subsection{How Does Policy Performance Change with Clutter?}
\label{sec:finding_clutter}

\begin{figure*}[t]
    \centering
    \begin{minipage}[t]{0.48\textwidth}
        \centering
        \includegraphics[width=\linewidth]{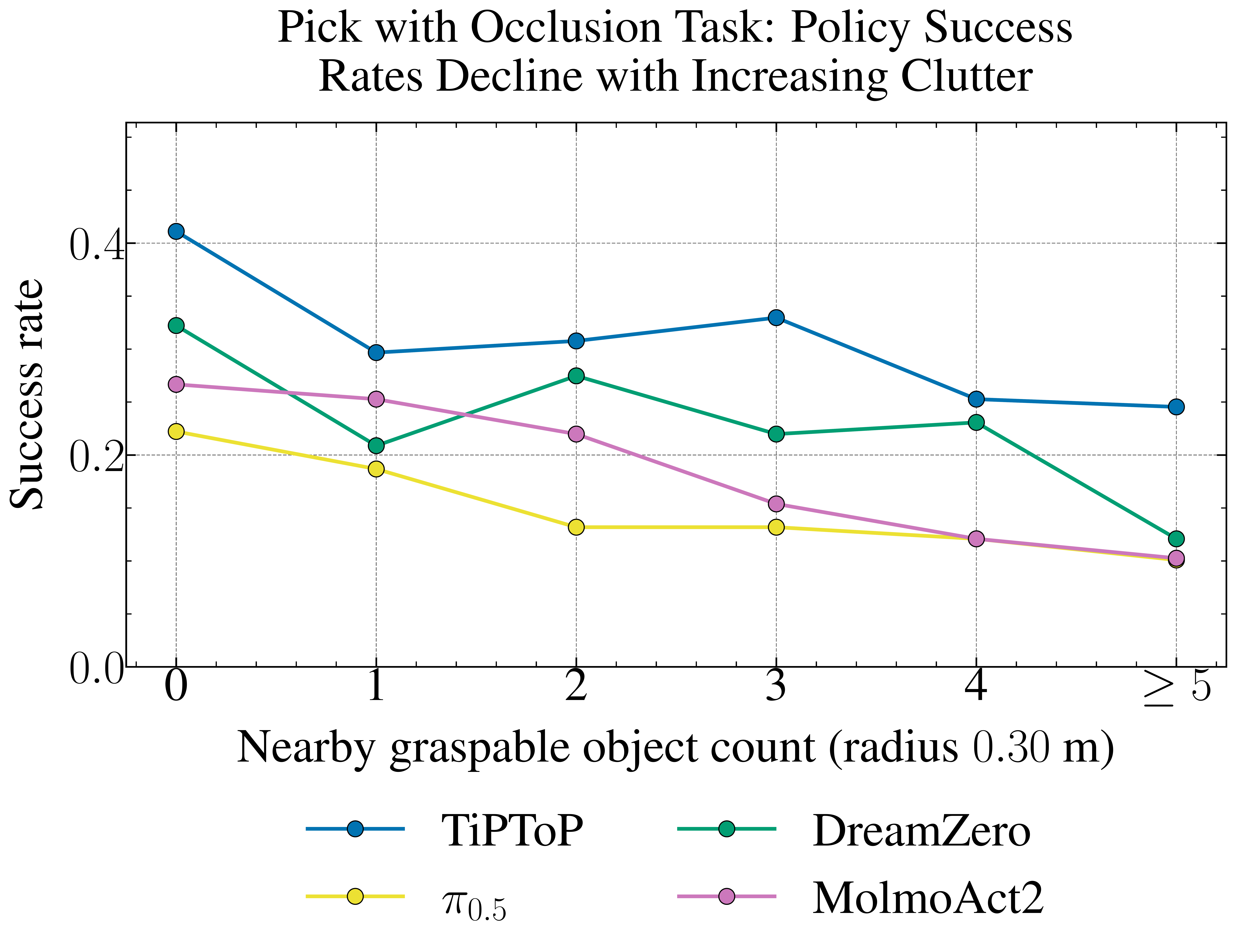}
    \end{minipage}\hfill
    \begin{minipage}[t]{0.48\textwidth}
        \centering
        \includegraphics[width=\linewidth]{figures/failure_modes_occlusion_density_delta.pdf}
    \end{minipage}
    \caption{Pick-with-occlusion task analysis. \textbf{(Left)} Policy success rates binned by the number of distractor objects within a small radius of the target pick object.
    \textbf{(Right)} Change in occurrence of failure modes as we go from sparse ($ \leq 1 $ nearby objects) to cluttered scenes ($ > 1 $  nearby objects)}
    \vspace{-1em}
    \label{fig:clutter_analysis}
\end{figure*}

In our physical reasoning tasks, success rates degrade consistently with scene clutter across all evaluated architectures. Fig.~\ref{fig:clutter_analysis} visualizes this in more detail for our Pick with Occlusion task, showing a near monotonic decline in success rate as the number of distractor objects near the target pick object increase. Additionally, success rates across all policies fall by over 50\% with 5+ distractor objects, relative to the zero-distractor setting. For this task, we additionally evaluate TiPToP~\cite{shen2026tiptopmodularopenvocabularyplanning}, a VLM-enabled modular planner (not a generalist policy). TiPToP reaches nearly double the performance of the best generalist policy, but also degrades significantly with clutter, indicating the difficulty is not specific to end-to-end policies.

Additionally, for a random sample of $ n = 200 $ scenes from the pick occlusion task, we evaluate different failure modes (see Fig.~\ref{fig:clutter_analysis}). After binning episodes by scene clutter (as measured by number of nearby distractor objects), we observe that in the high clutter subset, approaching/grasping the incorrect distractor object becomes a dominant failure mode, accounting almost entirely for the drop in success rate and indicating that perception is a weak link.

\begin{takeaway}
    Clutter significantly reduces current policies' performance across architectures, motivating data collection and evaluation in more complex environments going forward.
\end{takeaway}


\subsection{Are Policies Steerable by Language?}
\label{sec:finding_steering}

\begin{figure}[t]
    \centering
    \includegraphics[width=\columnwidth]{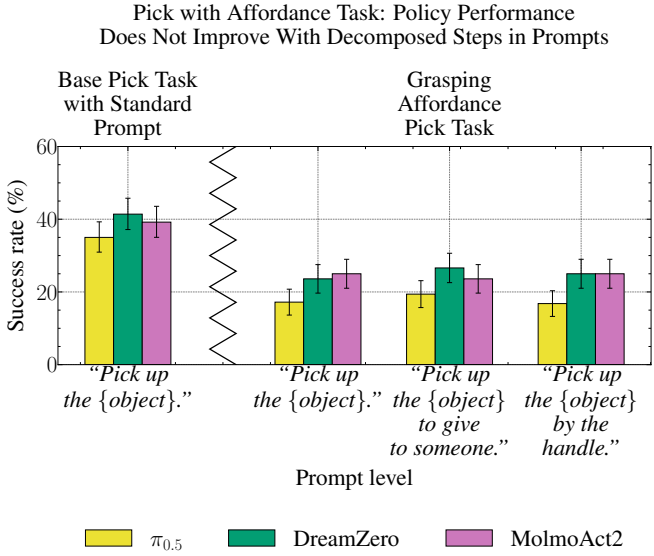}
    \caption{Success rates for base pick task (left), side by side with our Pick with Affordance task (right). We ablate multiple levels of prompt detail.}
    \label{fig:sempromptablation}
\end{figure}

On our Pick with Affordance task, we see that no policy exceeds a 30\% success rate. We address the possible explanation that these policies are simply poor at grasping in general by reporting success rates within the same environments but without the grasping location success condition (see Fig.~\ref{fig:sempromptablation}). Across all policies, success rates are nearly halved with the additional grasp location check. This points to failures specifically in reasoning about grasping affordances.

Additionally, we ablate the choice of prompt used in this task. Examples of our standard prompts include ``pick up the hot pan'' or ``pick up the mug to give to someone'', and we test both more specific instructions (``pick up the pan by the handle'') and the base task with no steering (``pick up the pan''). We find minimal impact from language steering across policies in either direction; neither added specificity nor its removal significantly shifts success rates.

\begin{takeaway}
    Policies show limited steerability with respect to specified affordances, suggesting that training datasets should pair precise language with the behaviors those instructions describe.
\end{takeaway}

\subsection{Sim-to-Real Correlation}

\begin{figure}[t]
    \centering
    \includegraphics[width=\columnwidth]{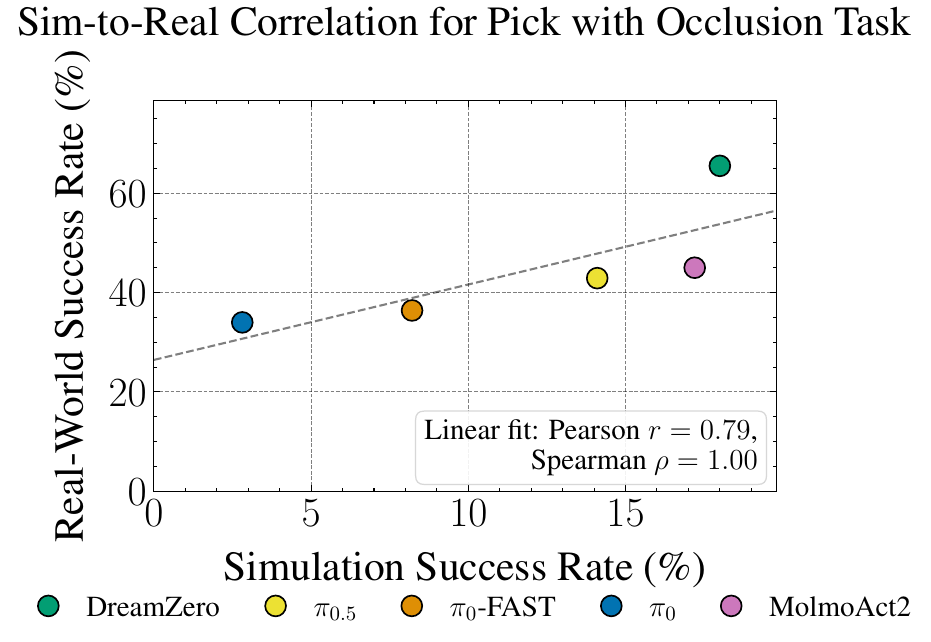}
    \caption{Sim-to-real correlation results for our Pick with Occlusion task.}
    \label{fig:sim_real_corr}
\end{figure}

Cluttered scenes are a significant new evaluation setting MolmoSpaces-Commonsense introduces over MolmoSpaces, which already establishes strong sim-to-real correlations for atomic skills (pick, open, close). To verify sim-to-real correlation in this regime, we run in-house real-world rollouts of MolmoAct 2 on pick-with-occlusion and supplement these with filtered episodes from RoboArena~\cite{atreya2025roboarenadistributedrealworldevaluation}, retaining medium-to-high clutter scenes --- 295 total real episodes against 1000 simulation episodes per policy. Rank order is preserved across the five evaluated policies (Fig.~\ref{fig:sim_real_corr}), with reasonable correlation ($r = 0.79$) between simulation and real success rates.

\subsection{Can VLMs Judge Policy Rollouts from MolmoSpaces-CommonSense?}
\label{sec:finding_vlm_judge}

\begin{table}[t]
\centering
\caption{VLM-as-judge accuracy (\%) for classifying rollout outcomes across three VLA policies ($ n = 200 $ rollouts per policy). Across the board, VLMs struggle to determine why a policy succeeded or failed in a given trajectory.}
\label{tab:vlm-judge-finegrained}
\begin{tabular}{lccc|c}
\toprule
Evaluator & $\pi_{0.5}$ & DreamZero & MolmoAct2 & Mean \\
\midrule
Molmo2                                  & 27.9          & 37.4          & 19.4          & 28.2 \\
Qwen3.5-122B-A10B                       & 44.6          & 46.1          & 44.8          & 45.2 \\
Gemini 3.1                              & \textbf{46.6} & \textbf{63.1} & 43.8          & 51.2 \\
Gemini Robotics-ER 1.6                  & 50.5          & 51.0          & \textbf{53.2} & \textbf{51.6} \\
\bottomrule
\end{tabular}
\end{table}

On the perception-only VLM judge task, no model exceeds a performance of $\sim$63\% (refer to Table~\ref{tab:vlm-judge-finegrained} for full results), indicating the perception and reasoning task alone is challenging and not just action prediction. We additionally notice common failure modes with rollout classification. In settings where the gripper does not make contact with any objects in the scene, models mistakenly output that the policy made contact and attempted to grasp an object (with some models, like Gemini 3.1, this happens more than 50\% of the time). For successful trajectories, VLMs often misclassify these as grasping failures (notably, Gemini Robotics ER 1.6 does this over 50\% of the time for DreamZero rollouts).

\begin{takeaway}
    VLMs struggle to judge trajectories, showing MolmoSpaces-CommonSense poses difficulties across the full stack of perception, reasoning, and action. This indicates that modular systems based on these models may still struggle on our tasks.
\end{takeaway}

\subsection{Do Policies Fail at Reasoning or at Execution?}

By design, MolmoSpaces-CommonSense defines tasks where reasoning and execution must be performed in tandem. We argue that including tasks isolating either reasoning or action would lead to a distribution of tasks not representative of real robotics applications. However, we still find that policies fail specifically at reasoning on our tasks, supporting that reasoning is required to attain 100\% success on these tasks.

A first example of policies failing at reasoning alone is on our Pick with Affordance task. As shown in Fig.~\ref{fig:sempromptablation}, we run an ablation of the tasks with a single variation: we omit the part-specific clause of the success condition. With this removed, the success rate approximately doubles, showing that all policies tested here are often capable of grasping. However, with the stricter part-specific success condition and with more specific prompts, policies are unable to grasp the object by the correct object part.

We additionally examine the Pick with Object Permanence task in more detail. As no openly
released end-to-end generalist policy that works zero-shot in our setting has memory (and therefore cannot solve this task by construction), we add a modular memory-enabled baseline: Gemini Robotics ER 1.6 points to the correct mug, then the Contact-Anchored Policy (CAP)~\cite{cui2026cap}
grasps it. We also add a random-pick baseline, which instead lifts a randomly
chosen mug with CAP, to illustrate the performance improvement that reasoning facilitates on this task.

\begin{table}[h]
\centering
\caption{Policy Performance on Pick with Object Permanence Task}
\footnotesize
\setlength{\tabcolsep}{3pt}
\refstepcounter{table}\label{tab:objpermanence-success}
\begin{tabular}{@{}lccc!{\vrule width 1pt}cc@{}}
\toprule
 & MolmoAct2 & $\pi_{0.5}$ & DreamZero & CAP$_{\mathrm{rand}}$ & CAP$_{\mathrm{mem}}$ \\
\midrule
SR (\%) & 0.00\,{\scriptsize$\pm$0.37} & 0.30\,{\scriptsize$\pm$0.57} & 0.30\,{\scriptsize$\pm$0.57} & 17.30\,{\scriptsize$\pm$2.47} & 28.20\,{\scriptsize$\pm$2.87} \\
\bottomrule
\vspace{-2em}
\end{tabular}
\end{table}

The memory-enabled policy improves over random-pick (28\% vs.\ 17\%, Tab.~\ref{tab:objpermanence-success}), approaching the $2\times$ gain perfect mug identification would predict with two candidate mugs; this gap stems specifically from memory and reasoning capability.

\subsection{Is Architecture or Data the Bottleneck to Performance on These Tasks?}

Across our experiments, architecturally distinct policies (VLA, VLA with Reasoning, WAM) exhibit similar failure patterns. On our Pick with Occlusion task, we analyze the correlation between $\pi_{0.5}$, MolmoAct2, and DreamZero's successes. We run a $ \chi^2 $ goodness-of-fit test against a null of mutual independence (each policy succeeding at its own rate). Given 1k Pick-with-Occlusion environments, we compute the expected vs. actual episode count for each of the 8 possible joint outcomes, finding $ \chi^2 = 264.3 $, $ p < 1 \times 10^{-55} $ (d.f. = 4). Independence predicts 5.5 episodes where all three succeed; we observe 38. The episode difficulty is evidently shared across architecture classes.

We next investigate the role of training data in the results shown on the Pick-with-Occlusion task. Each of the policies tested was finetuned on the DROID dataset. We use Fable 5.1~\cite{anthropic2026fable} to annotate 396 random DROID episodes with their degree of clutter. Specifically, we process the RGB images and identify the number of objects within 1.5 target-widths of the target object (measured in pixel space). Though DROID scenes are reasonably populated, they offer limited clutter around the target object (Tab.~\ref{tab:droid_clutter}); only around 18\% of the dataset includes scenes with 4+ objects near the target. Data collection in more densely cluttered scenes is likely to be important in the future. Instructions also almost never reference a hidden target or specific grasp locations. We therefore hypothesize that finetuning data limitations, rather than architectural differences, account for most of the observed failures.

\begin{table}[h]
\centering
\caption{Distribution of clutter around the target object in DROID.}
\label{tab:droid_clutter}
\begin{tabular}{ccc}
\toprule
\textbf{\# Nearby Objects} & \textbf{Episodes} & \textbf{\% (95\% CI)} \\
\midrule
0    & 54  & 13.6 (10.6--17.4) \\
1    & 109 & 27.5 (23.4--32.1) \\
2    & 88  & 22.2 (18.4--26.6) \\
3    & 76  & 19.2 (15.6--23.4) \\
4    & 32  & 8.1 (5.8--11.2) \\
$5+$ & 37  & 9.3 (6.9--12.6) \\
\bottomrule
\vspace{-2em}
\end{tabular}
\end{table}

\vspace{-1.5em}


\section{Discussion}
\label{sec:limitations}

A notable pattern in our analysis is that architecturally distinct policies (VLAs, WAMs) fail in highly correlated ways: all evaluated policies deteriorate similarly under clutter, with wrong-object selection as the dominant failure mode, and show shared insensitivity to language steering. This pattern suggests that gaps exist upstream of architecture, in the existing training datasets and evaluation procedures used for robot policies, and we hope that the simulation environments in MolmoSpaces-CommonSense can be used to address these gaps. Our analysis also shows that the difficulty of tasks in our commonsense taxonomy extends to all parts of the perception-reasoning-action stack.

\noindent\textbf{Limitations and Future Work.}
We focus on evaluating policies zero-shot, which limits our scope of analysis to the subset of our taxonomy on which these policies attain a threshold success. In future work, we plan to leverage these environments to train policies, analyzing existing shortcomings and uncovering improvements for commonsense reasoning. As generalist policies improve, we also hope to augment our set of tasks to cover more advanced capabilities within our commonsense taxonomy (e.g.\ higher stages of object permanence understanding), as well as additional categories of commonsense reasoning (e.g.\ intuitive physics).
We additionally caveat our VLM-judge experiments: the judge is a model distinct from the policies being evaluated, so it bounds rather than pinpoints the specific reasoning failures of the policies themselves. We treat the experiment as analytically useful nonetheless --- verifying a rollout draws on perceptual and reasoning capabilities similar to those required to solve the task.

\noindent\textbf{Conclusion.}
We present MolmoSpaces-CommonSense, a benchmark that tests the combination of commonsense reasoning and action for generalist policies. Our goal is to facilitate a new line of inquiry into building robot policies that can not only execute atomic skills but also reason through the commonsense-laden situations the real world presents.

\section{Acknowledgements}
\label{sec:acknowledgements}

We thank Winson Han for valuable advice on figure design and Donovan
Clay for support with the real-world experiments. We also thank Omar
Rayyan for support with the CAP evaluations and Jordi Salvador for help packaging the benchmark assets for the code release. AG is supported by the NSF CSGrad4US Fellowship. Additionally, we thank Ai2 and the UW Hyak and Tillicum high-performance computing clusters for providing us with compute resources.

\newpage




\bibliographystyle{IEEEtran}   
\bibliography{main}

\clearpage

\appendix
\label{sec:appendix}

Below, we include additional details on per-task implementation and sampling, policy parameters for evaluation, and VLM usage.

\subsection{Per-Task Implementation Details}

We elaborate on simulation implementation details per commonsense taxonomy task here. First, in Table~\ref{tab:task-horizons}, we share the horizon used per task. Per-task, we also describe environment sampling and task implementation (including details like success condition, etc.).

\begin{table}[H]
\centering
\caption{Maximum episode horizon per task. Episodes that reach the horizon without satisfying the success condition are marked as failures. All evaluated policies are run with a policy $\Delta t$ of 66 ms.}
\label{tab:task-horizons}
\begin{tabular}{lc}
\toprule
Task & Horizon (Policy Steps) \\
\midrule
Pick with Occlusion          & 300  \\
Pick with Covering           & 300  \\
Pick with Containment        & 500  \\
Pack with Occlusion          & 1500 \\
Pack with Covering           & 1500 \\
Pick with Affordance         & 300  \\
Stack with Affordance        & 600  \\
Pick with Object Permanence  & 300  \\
\bottomrule
\end{tabular}
\end{table}

\textbf{Pick with Occlusion, Pick with Covering}

These environments use MSProc scenes. Scenes are updated with a procedural cluttering step. In this step, all graspable objects within a filtered superset are moved to within a radius of the target object, placed such that no contact exists at the time of placement (via MuJoCo contact checks), and settled.

Between occlusion and covering tasks, the only difference is the parameters used for procedural cluttering. Clutter object placement uses a spherical coordinate system with the task's target pick object as its origin. The primary parameter varied for covering is the polar angle $ \phi $- a lower value is used for covering, denoting higher z values (and closer to the target object, given a fixed radius $ r $); this achieves clutter object positions on top of and covering the target pick object. In contrast, higher $ \phi $ in the occlusion setting typically achieves clutter objects surrounding but not covering the target pick object.

For both tasks, the success condition is simply a successful lift of the target object (via object z coordinate), combined with a contact check between the gripper and the object.

\textbf{Pick with Containment}

These environments use MSCrafted scenes. Scenes are sampled with a target drawer (articulated object) as well as a target pick object. A position is sampled for the target pick object inside the closed drawer, using the drawer handle as a reference point.

The task success condition has 2 stages. The first stage is the successful opening of the drawer (measuring the degree of articulation of the sliding joint). The second stage is the successful pick of the object (see above for the pick task success condition).

\textbf{Pack with Occlusion, Pack with Covering}

These environments use MSProc scenes. The scene initially includes a single target object, along with an added box receptacle for placement. Scenes then use the same procedural cluttering additional step as above (i.e., same occlusion/covering split) to introduce more target objects for packing. Robot placement during task sampling is done such that both the receptacle and pick objects lie within its workspace.

The success condition of packing tasks is the placement of the newly placed clutter objects and the originally present object into the box. For each object, we do a cascading set of checks to verify box containment. The first check is a contact check with the box, which additionally requires the contact normal to be facing up (dot product threshold with positive $ z $ axis in the box's coordinate frame). This check may fail due to either improperly registered MuJoCo contacts or items being stacked in the box. Therefore, we do an additional geometric check, where rays are cast outward from the object origin along the box $ \pm x $ and $ \pm y $ axes. If all rays eventually intersect with the box geometry, we consider the object as packed. The packing task must have all objects satisfy the packed condition to be marked as successful.

\textbf{Pick with Affordance}

These environments use MSCrafted scenes, with no additional modifications. The key implementation detail with this task lies in its success condition. We leverage the already generated antipodal grasps per object within the MolmoSpaces ecosystem, and annotate the grasps per object for correctness given the task of picking the object by the handle (or picking up the utensil to give to someone, picking up the hot pan, etc.).

\begin{figure}[t]
    \centering
    \includegraphics[width=\columnwidth]{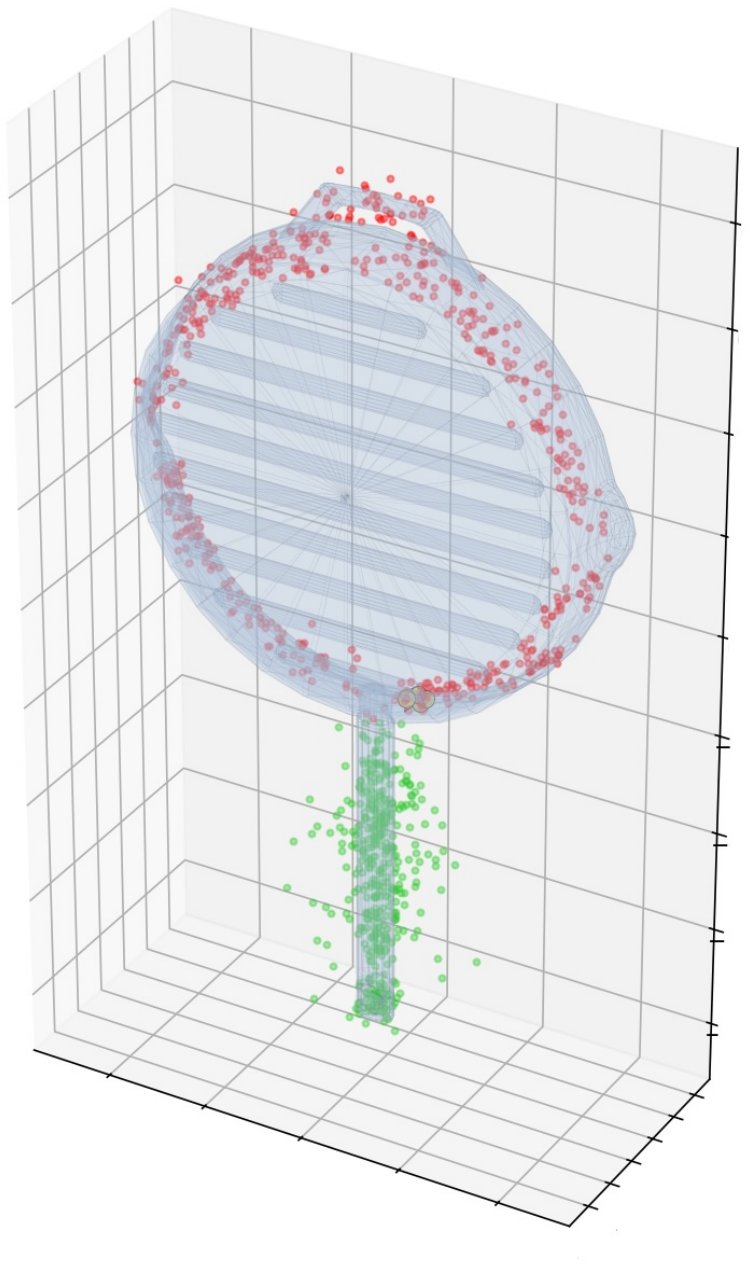}
    \caption{Grasps classified as good and bad using our grasp annotation tool, given the task ``pick up the hot pan.'' A semantically correct grasp given this command requires picking up the pan by the handle.}
    \label{fig:pangrasps}
\end{figure}

Within the task success condition, all grasps for the object are loaded. If the object is successfully lifted and the gripper has made contact, a nearest neighbors check is invoked against the set of grasps. Our distance metric is simply the translation-only distance between the fingertip-midpoint tool center point and a generated grasp's translation component. We consider the 5 nearest grasps, and if $ \geq 3 $ of those lie within the set of grasps classified as ``good'' given the task, success is returned. Fig.~\ref{fig:pangrasps}  shows an example of the grasp annotations for the task ``pick up the hot pan''. The nearest neighbors to the current grasp must be those in green (i.e., on the pan's handle) for the task to be a success.

\textbf{Stack with Affordance}

Environments use MSCrafted scenes. We additionally place 2-4 blocks within the robot's workspace in the scene. All blocks are sized differently, with size options including 2 cm, 4 cm, 6 cm, and 8 cm side lengths.

The success condition requires that blocks are stacked in order of size (with the largest on the bottom and the smallest on top). Note that block sizes are sufficiently different such that this is generally required for any stacking success. Concretely, for every possible pair of larger block and smaller block, we verify by both a contact check and a fallback geometric check if the smaller block is immediately above the larger block. If this holds for $ n_{\text{blocks}} - 1 $ pairs, then our success condition returns True.

\textbf{Pick with Object Permanence}

Environments use MSProc scenes. We additionally place 2 upside-down mugs in midair above a surface in the scene, with a ball on the surface immediately underneath one of the mugs. At the first task frame, the mugs are still suspended in midair, which allows policies a window of observing the ball prior to the scene settling and the mugs falling to cover the ball. The initial visibility of the ball is verified by a segmentation mask check on the ball in the first frame of the task.

For the first 5 seconds of simulation time, policy outputs are set to no-operation. In other words, the policy is forced not to act prior to the scene settling and the mugs falling to obscure the ball. The policy still receives observations during this time (necessary for solving the task).
From here, the task's success condition is a simple pick task success condition (see above), with the target object being the mug covering the ball.

\subsection{Planner for Task Feasibility Verification}

In our task sampling process, after sampling environments, we run a planner-based policy on each. We filter to environments where the planner policy succeeds in order to arrive at a candidate set of environments, from which the final 500-1000 environments per task are chosen. Note that the 100\% planner success rate on our tasks in Table~\ref{tab:headline_results} follows by construction from this procedure.

Our planner policy follows a procedure similar to MolmoBot's data generation (for manipulation tasks with DROID). We use the already generated grasps for the target object, sample one of the grasps, and then generate pregrasp, grasp, preplace, and place poses as needed. We then use inverse kinematics to execute this sequence of gripper poses (i.e., no collision avoidance, etc.).

In clutter tasks, this ends up being a sub-optimal way to solve tasks; the gripper may collide with and knock over objects on its way to the target object. However, it at least verifies that the task is solvable. In packing tasks, we attempt to minimize the planner's suboptimality by sequencing the multiple pick and places such that the planner picks clutter objects in reverse order of their placement during cluttering. In a covering setting, for example, the planner would pack the top item in a pile first, before packing the items below.

For the Pick with Containment task, our planner has 2 stages. The first stage generates poses for grasping and pulling the drawer handle, and the second stage generates poses for picking up the object within the drawer.

\subsection{Policy Parameters}

\begin{table*}[t]
\centering
\caption{Per-policy evaluation configurations.}
\label{tab:policy_eval_configs}
\small
\setlength{\tabcolsep}{4pt}
\renewcommand{\arraystretch}{1.15}
\begin{tabularx}{\textwidth}{@{}l XXXX@{}}
\toprule
\textbf{Parameter} & \textbf{MolmoAct2} & \textbf{$\pi_0$, $\pi_{0}$-FAST, $\pi_{0.5}$} & \textbf{DreamZero} & \textbf{TiPToP} \\
\midrule
Policy Type & Learned & Learned & Learned & TAMP (Planner) \\
Finetuning & DROID & DROID & DROID & N/A \\
Output Format & Joint Position & Joint Position & Joint Position & Joint Position \\
Policy $\Delta t$ & 66 ms / 15.15 Hz & 66 ms / 15.15 Hz & 66 ms / 15.15 Hz & 66 ms / 15.15 Hz \\
Actions Executed per Chunk & 8 & 8 & 24 & N/A, Open-Loop \\
Cameras & Wrist + Exo & Wrist + Exo & Wrist + Exo & Wrist + Exo, + Depth \\
Input Image Resolution & 640$\times$480 (Full Render) & 224$\times$224 (Resize w/ Pad) & 180$\times$320 (Resize w/ Pad) & 640$\times$480 (Full Render) \\
Gripper Decode & Binary @ 0.5 & Binary @ 0.5 & Binary @ 0.5 & Binary @ 0.5 \\
End on Success & True & True & True & False \\
\bottomrule
\end{tabularx}
\end{table*}

Table~\ref{tab:policy_eval_configs} shows details on key parameters used per policy in evals. Note that all learned policies tested are joint position policies finetuned on DROID; we use publicly released DROID checkpoints for all. We additionally binarize the gripper output, thresholding at the midpoint of the range of gripper values. We execute the first $k$ actions of each predicted chunk (where $ k $ is the per-policy value in Table~\ref{tab:policy_eval_configs}) before re-querying the policy. For all learned policies, we terminate the task upon success. For TiPToP, this is unnecessary since the planner executes an open loop before stopping. For more details on all relevant parameters, refer to the table above.

\subsection{More Details on VLM Evaluations}

The VLM-judge ground truth consists of Pick with Occlusion policy rollouts annotated by the authors: we sampled approximately 200 episodes per policy, yielding final counts of 204 ($\pi_{0.5}$), 206 (DreamZero), and 201 (MolmoAct2), for $ n = 611 $ total. All accuracies in Tables~\ref{tab:vlm-judge-finegrained} and~\ref{tab:vlm-judge-finegrained-full} and the confusion matrices in Tables~\ref{tab:gemini31-aggregated-cm} and~\ref{tab:gemini-robotics-er-aggregated-cm} are computed over this set. This annotated set also underlies the failure-mode analysis of Sec.~\ref{sec:finding_clutter}, where we restrict to the $n = 201$ environments shared across all three policies.

We now elaborate on the classifications of failure modes (or success) used for these evaluations:

\begin{itemize}
    \item \textbf{No Grasp Attempt} - the gripper does not come into contact with any graspable object in the scene
    \item \textbf{Wrong Object} - the gripper comes into contact and/or successfully grasps a different object from the target
    \item \textbf{Success} - the correct object is successfully picked up
    \item \textbf{Clutter Collision} - the robot attempts the correct object but fails due to a collision with a clutter object
    \item \textbf{Grasping Failure} - the robot attempts the correct object but fails to grasp it after making contact
\end{itemize}

Table~\ref{tab:vlm-judge-finegrained-full} expands Table~\ref{tab:vlm-judge-finegrained} with all evaluated configurations; Table~\ref{tab:vlm-judge-finegrained} reports the best configuration per model family by mean accuracy (Gemini 3.1 low, Gemini Robotics-ER 1.6 low, Qwen3.5-122B-A10B). Interestingly, setting Gemini models' thinking budget to higher values did not improve performance.

\begin{table}[H]
\centering
\caption{VLM-as-judge fine-grained accuracy (\%) across rollouts from three VLA policies. Best per column in \textbf{bold}. Models without bibliography entries: GPT-5.4 via the OpenAI API
(\texttt{gpt-5.4}, accessed 2026-05-28); Qwen3-VL-8B
(\url{https://huggingface.co/Qwen/Qwen3-VL-8B-Instruct}); Qwen3.6-35B-A3B
(\url{https://huggingface.co/Qwen/Qwen3.6-35B-A3B}).}
\label{tab:vlm-judge-finegrained-full}
\resizebox{\columnwidth}{!}{%
\begin{tabular}{lcccc}
\toprule
Evaluator & $\pi_{0.5}$ & DreamZero & MolmoAct2 & Mean \\
\midrule
Gemini 3.1 (high)                       & 45.1          & 52.9          & 45.8          & 47.9 \\
Gemini 3.1 (low)                        & 46.6          & \textbf{63.1} & 43.8          & 51.2 \\
Gemini 3.1 (medium)                     & 44.6          & 53.9          & 42.8          & 47.1 \\
Gemini Robotics-ER 1.6 (high)           & \textbf{51.0} & 50.5          & 52.2          & 51.2 \\
Gemini Robotics-ER 1.6 (low)            & 50.5          & 51.0          & \textbf{53.2} & \textbf{51.6} \\
GPT-5.4 (high)                          & 45.1          & 46.1          & 46.8          & 46.0 \\
Qwen3-VL-8B                             & 26.5          & 30.6          & 47.3          & 34.8 \\
Qwen3.5-35B-A3B (reasoning)             & 38.9          & 44.1          & 41.1          & 41.4 \\
Qwen3.6-35B-A3B (reasoning)             & 42.0          & 46.5          & 34.8          & 41.1 \\
Qwen3.5-122B-A10B                       & 44.6          & 46.1          & 44.8          & 45.2 \\
Qwen3.5-397B-A17B (fp8)                 & 41.1          & 48.1          & 36.8          & 42.0 \\
Molmo2                                  & 27.9          & 37.4          & 19.4          & 28.2 \\
\bottomrule
\end{tabular}%
}
\end{table}

We also share confusion matrices that elaborate on the points mentioned in Section~\ref{sec:finding_vlm_judge}. In Table~\ref{tab:gemini31-aggregated-cm}, note that in rollout videos where the policy does not attempt to grasp anything (i.e. makes contact with no graspable object in the scene), more than 50\% of the time the model outputs that an object has been attempted and a different failure mode occurs. Table~\ref{tab:gemini-robotics-er-aggregated-cm} shows that Gemini ER 1.6 misclassifies successful trajectories as grasping failures a majority of the time. For more details on the various failure modes, refer to the confusion matrices below.

\begin{table}[H]
\centering
\caption{Aggregated confusion matrix for Gemini 3.1 (low reasoning) judged across all rollout sources (DreamZero, $\pi_{0.5}$, MolmoAct2; $n{=}611$). Rows are ground-truth labels, columns are model predictions; diagonal entries (correct predictions) are in \textbf{bold}.}
\label{tab:gemini31-aggregated-cm}
\small
\setlength{\tabcolsep}{4pt}
\resizebox{\columnwidth}{!}{%
\begin{tabular}{lcccccc}
\toprule
True $\backslash$ Pred & No Grasp & Wrong & Success & Clutter & Grasping & Total \\
 &  Attempt &  Object & Success &  Collision &  Failure &  \\

\midrule
No Grasp Attempt   & \textbf{87} & 59           & 22          & 2           & 7            & 177 \\
Wrong Object       & 15          & \textbf{126} & 10          & 3           & 3            & 157 \\
Success            & 1           & 8            & \textbf{77} & 1           & 5            & 92  \\
Clutter Collision  & 0           & 15           & 10          & \textbf{1}  & 0            & 26  \\
Grasping Failure   & 11          & 49           & 74          & 3           & \textbf{22}  & 159 \\
\midrule
Total              & 114         & 257          & 193         & 10          & 37           & 611 \\
\bottomrule
\end{tabular}%
}
\end{table}

\begin{table}[H]
\centering
\caption{Aggregated confusion matrix for Gemini Robotics-ER 1.6 (low reasoning) judged across all rollout sources (DreamZero, $\pi_{0.5}$, MolmoAct2; $n{=}611$). Rows are ground-truth labels, columns are model predictions; diagonal entries (correct predictions) are in \textbf{bold}.}
\label{tab:gemini-robotics-er-aggregated-cm}
\small
\setlength{\tabcolsep}{4pt}
\resizebox{\columnwidth}{!}{%
\begin{tabular}{lcccccc}
\toprule
True $\backslash$ Pred & No Grasp & Wrong & Success & Clutter & Grasping & Total \\
 &  Attempt &  Object & Success &  Collision &  Failure &  \\
\midrule
No Grasp Attempt   & \textbf{98}  & 43           & 0           & 1           & 35           & 177 \\
Wrong Object       & 31           & \textbf{104} & 1           & 1           & 20           & 157 \\
Success            & 7            & 9            & \textbf{12} & 2           & 62           & 92  \\
Clutter Collision  & 2            & 10           & 0           & \textbf{3}  & 11           & 26  \\
Grasping Failure   & 23           & 25           & 8           & 5           & \textbf{98}  & 159 \\
\midrule
Total              & 161          & 191          & 21          & 12          & 226          & 611 \\
\bottomrule
\end{tabular}%
}
\end{table}

Below, we share the prompt used for all VLM evals. Each query consists of 9 images: a goal image (an early frame with the target object boxed in red, plus a zoomed-in crop) followed by 8 frames sampled uniformly from the rollout. We parse the predicted class from the VLM inference output. The prompt's output labels map to our categories as: Grasping success → Success; Grasping failure, correct object → Grasping Failure; Grasping failure, clutter → Clutter Collision; Grasping failure, wrong object → Wrong Object; No grasp attempt → No Grasp Attempt.


\begin{figure*}[!t]
\noindent VLM-as-judge Evaluation Prompt:
{\footnotesize
\begin{verbatim}
I want you to evaluate the performance of a vision language action
model on the given robotics task (a cluttered scene task in a
MuJoCo-based simulation benchmark). The task of the robot is to pick
up a particular object within the cluttered scene. I will pass in 9
images relevant to the task and the robot's policy rollout. The first
image depicts the target object that needs to be picked up. It
includes a frame early in the task, with the target object boxed in
red. That image also includes a zoomed in crop of the target object.
The next 8 images shown are in chronological order and come from the
robot executing the task. Carefully analyze these sequential images
and classify the success or failure mode that the policy experiences-
i.e. whether the robot is successful in picking up the target object
specified in the first image, based on how the rollout progresses in
the last 8 images. Use the following partition of the outcome space
for this task:

The robot doesn't attempt to grasp anything- the black robot gripper
does not make contact with any object in the scene, only empty
surfaces or nothing at all

Attempts to grasp an object in the scene ->
a. Does it go for the wrong object- this requires the black robot
   gripper making contact with an object OTHER THAN THE TARGET OBJECT
   specified in the first image of the goal object boxed, but does
   not require a good grasp
b. Does it go for the right object- this requires the black robot
   gripper making contact with the target object specified in the
   first goal image.
   i.  Does it successfully grasp the object- this requires the
       gripper closing around the object
   ii. Does it fail- the gripper does not successfully close around
       the object, but it still makes contact with the object
       1) Does it fail due to clutter- As the black robot gripper
          moves toward the target object, does it run into another
          object along the way?
       2) Does it just have a grasping failure, where the gripper
          approaches and makes contact with the correct object, but
          fails to close around the object

In addition to your reasoning, you should provide a final
classification of this episode as one of the below, given the above
partition of the outcome space:

Grasping success
Grasping failure, correct object
Grasping failure, clutter
Grasping failure, wrong object
No grasp attempt

Do your diagnosis after carefully considering these factors:
- Always be aware of which object is the target. For this, refer to
  the first image passed in, and the object boxed in red and with a
  zoomed in crop.
- The path the policy takes. Does it run into other objects on its
  path toward the target object
- Does the robot stop and attempt to grasp other objects, or does it
  keep going towards the target object
- Closely examine the black robot gripper in all frames. What is it
  holding, if it is holding anything? Does the held item match the
  target item boxed in red on the right side of each image? Is it
  fully closed around the object? What is the gripper making contact
  with, if it is making contact with anything?
\end{verbatim}
}
\end{figure*}

\end{document}